\documentclass{article}
\usepackage{spconf}
\usepackage{microtype}      
\usepackage{amsmath,amssymb,amsthm,mathrsfs,amsfonts,dsfont,bm} 
\usepackage{booktabs}
\usepackage{nicefrac}
\usepackage{algorithm}
\usepackage{algpseudocode}
\usepackage{color}

\usepackage{times}
\usepackage{helvet}
\usepackage{courier}
\usepackage{graphicx}
\usepackage{subcaption}
\ninept

\newcommand{\Lo}{\mathcal{L}}

\newcommand{\E}{\mathds{E}}
\newcommand{\X}{\mathcal{X}}
\newcommand{\C}{\mathcal{C}}
\newcommand{\Y}{\mathcal{Y}}

\newcommand{\var}{\mathds{V}\mathrm{ar}}

\newcommand{\la}{\ell}

\newcommand{\Z}{\mathcal{Z}}

\newcommand{\dz}{\,\mathrm{d}z}

\newcommand{\dy}{\,\mathrm{d}y}

\newcommand{\w}{\mathbf{w}}

\title{SHADE: Information-Based Regularization for Deep Learning}
\name{Michael Blot$^1$, Thomas Robert$^{1}$,  Nicolas Thome$^2$, Matthieu Cord$^1$}
\address{$^1$ Sorbonne Universit{\'e}s, UPMC Univ Paris 06, CNRS, LIP6 UMR 7606, 4 place Jussieu, 75005 Paris \\
        $^2$ CEDRIC, Conservatoire National des Arts et M\'etiers, 292 Rue St Martin, 75003 Paris, France \\
        \small \{michael.blot, thomas.robert, matthieu.cord\}@lip6.fr - nicolas.thome@cnam.fr}

\begin{document}
%
\maketitle
\begin{abstract}
Regularization is a big issue for training deep neural networks. In this paper, we propose a new information-theory-based regularization scheme named SHADE for SHAnnon DEcay. The originality of the approach is to define a prior based on conditional entropy, which explicitly decouples the learning of invariant representations in the regularizer and the learning of correlations between inputs and labels in the data fitting term. Our second contribution is to derive a stochastic version of the regularizer compatible with deep learning, resulting in a tractable training scheme. We empirically validate the efficiency of our approach to improve classification performances compared to standard regularization schemes on several standard architectures.
\end{abstract}

\begin{keywords}
Deep learning, regularization, invariance, information theory, image understanding
\end{keywords}

\section{Introduction}
\label{sec:intro}

    Deep neural networks (DNNs) have shown impressive state-of-the-art results in the last years on numerous tasks and especially for image classification \cite{alexnet,resnet}. One key element is the use of very deep models with a huge number of parameters. Accordingly, DNNs need to be trained on a lot of data (\textit{e.g.} ImageNet) and with a regularization scheme to control overfitting.
    Although regularization methods such as weight decay \cite{weightdecay}, dropout \cite{dropout} or batch normalization \cite{batchnorm} are common practice, the question of DNN regularization remains open as demonstrated by \cite{rethinking}.
    
    
    
    Formally, let us note $X \in \X$ the input variable, $C \in \C$ the output (class) variable, $w$  model parameters and $Y=h(w,X)$ the (deep) representation of the input that leads to the class prediction.
    Usually, training schemes for deep models for classification tasks use an objective function  which 
    linearly combines a classification loss  $\ell_{\mathrm{cls}}(w, X, Y, C)$ --\,generally cross-entropy\,--  and a regularization term $\Omega(w, X, Y,C)$, with $\beta \in \mathbb{R}^+$:
    \begin{equation}
    \label{eq:loss}
        \Lo(w) = \E_{(X,C)} \big( \ell_{\mathrm{cls}}(w, X, Y, C) + \beta\cdot\Omega(w, X, Y, C) \big)
    \end{equation}
    


 This paper studies the issue of regularization, and we propose a new regularization term  $\Omega(w, X,Y, C)$ in Eq~(\ref{eq:loss}).

   
    \paragraph*{Quantifying invariance.} Designing DNN models that are robust to variations on the training data and that preserve class information is the main motivation of this work. 
    With this motivation, Scattering decompositions \cite{scattering} are appealing transforms, which have been incorporated into adapted network architectures like \cite{bruna}.
    However, for tasks like image recognition, it is very difficult to design an explicit modelling of all transformations a model should be invariant to. 

    \paragraph*{Information-theory-based regularization.}
    Many works like \cite{DBLP:journals/corr/PereyraTCKH17} use information measures as regularization criterion. 
    The Information Bottleneck framework (IB) proposed in \cite{IB} suggests to use mutual information $I(X,Y)$ (see \cite{element} for definition) 
    as regularization criterion. \cite{IBvariational} extends it to a variational context VIB. 
    However, regularization based on $I(X,Y)$ may conflict with the task loss $\ell_{\mathrm{cls}}(w, X, C)$ in Eq~(\ref{eq:loss}). In addition, IB \cite{IB}  is computationally expensive and is only applied at a single final layer of the network.
     


    In this paper, we propose a new regularization method, denoted as SHADE for SHAnnon DEcay. Our first contribution is to design a new regularization loss, that aims at minimizing a particular criterion: the entropy of the representation variable conditionally to the class variable, \textit{i.e.} $\Omega(w, X, C)=H(Y\mid C)$. This criterion strongly supports intra-class invariance of the representation, without conflicting with $\ell_{\mathrm{cls}}(w, X, C)$ in Eq~(\ref{eq:loss}). Our second contribution consists in deriving a tractable surrogate function of $H(Y\mid C)$. This enables the incorporation of the regularizer at every layer of the network, leading to a scalable optimization scheme based on stochastic gradient descent (SGD). 

    We provide an extensive experimental validation of our SHADE regularizer for important standard DNNs, namely AlexNet, ResNet and Inception, applied to CIFAR-10 and ImageNet datasets.

\section{SHADE: A new Regularization Method}
\label{sec:shade}

    In this section, we further describe SHADE, a new regularization term based on the conditional entropy $H(Y\mid C)$ designed to drive the optimization towards a more invariant representation.

    \subsection{Conditional Entropy-based Regularization for Deep Neural Networks}
  
\begin{figure}[tbp]
                 \includegraphics[width=0.49\textwidth]{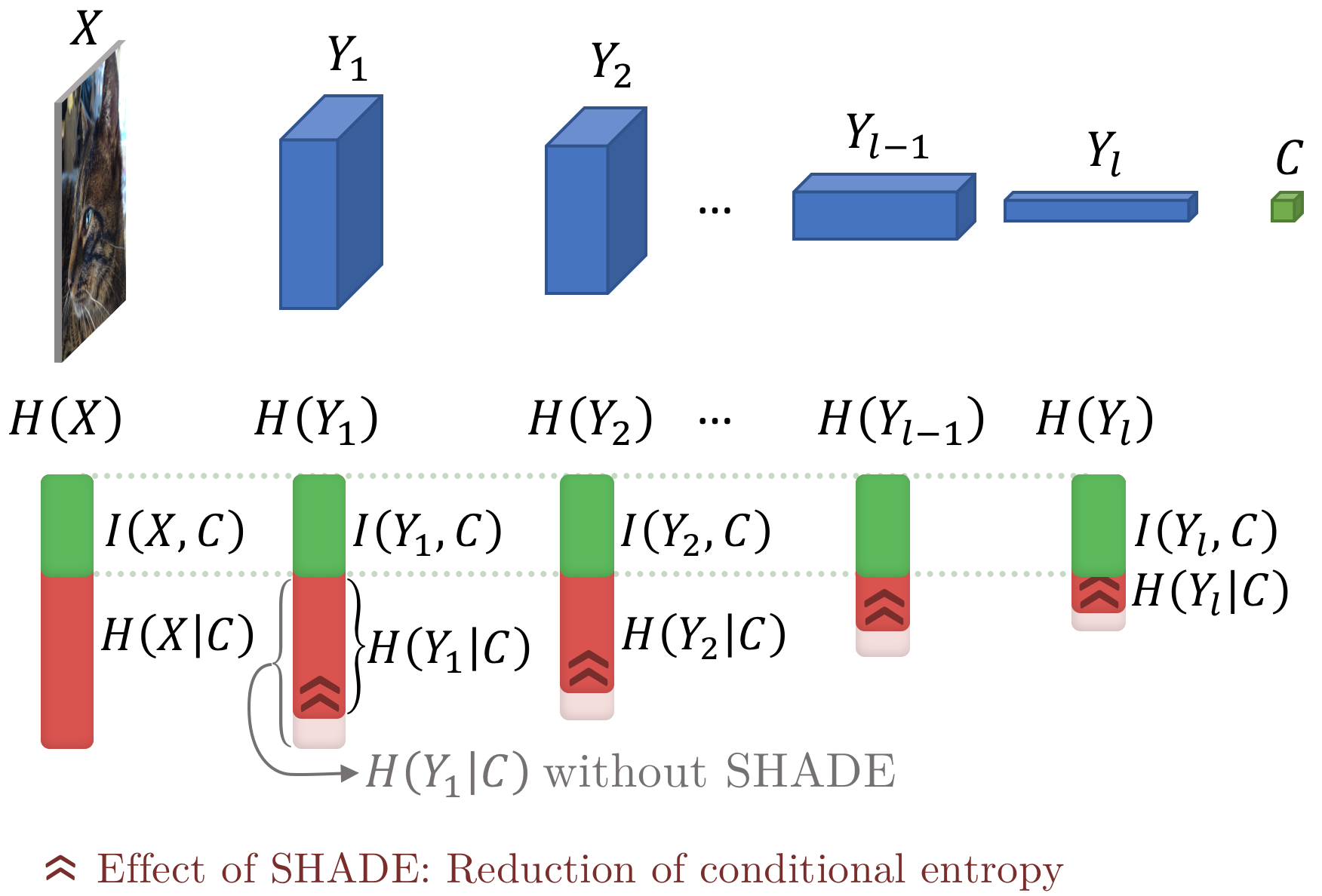}
            \caption{DNN architecture with corresponding layers' entropies, showing the layer-wise action of SHADE. Given that $H(Y_i) = I(Y_i,C)+H(Y_i\mid C)$, SHADE minimizes $H(Y_i\mid C)$ without affecting $I(Y_i,C)$.}
            \label{fig:layer-wize-regularization}
        \end{figure}

    In this article, the considered task is the classification of images, so we focus on intra-class invariance, explaining the use of $H(Y\mid C)$ as a criterion. 
    An overview of the approach is given in Fig.~\ref{fig:layer-wize-regularization}.

        Our approach differs from the Information Bottleneck framework (IB) \cite{IB}, which suggests to use $I(X,Y)$ as a regularizer. In the case where $\X$ is discrete, our criterion is related to IB's through the following development $I(X, Y) = I(C, Y) + H(Y\mid C)$ that holds for deterministic models ($H(Y\mid X)=0$). In a context of optimization with SGD, minimizing $H(Y\mid C)$ appears to be more efficient to preserve the term $I(C, Y)$, which represents the mutual information of the representation variable with the class variable and must stay high to predict accurately $C$ from $Y$. It is illustrated in Fig.~\ref{fig:layer-wize-regularization} where we see that $I(Y_l,C)$ (in green) remains constant while $H(Y_l\mid C)$ (in red) decreases.
        
        We claim that {$H(Y\mid C)$} quantifies accurately how invariant a representation is, while being agnostic to the transformations it is invariant to. When developing the entropy of the representation, we get $H(Y) = I(X,Y) + H(Y \mid X)$. Given that the DNN is deterministic, we have $H(Y\mid X)=0$ so $H(Y) = I(X, Y) = H(X) - H(X \mid Y)$. $H(X)$ being fixed, $H(Y)$ is inversely related to $H(X\mid Y)$. This last term can lower bound any input reconstruction error, demonstrated with inequalities such as Fano's inequality \cite{element}. Thus, it can perfectly quantify how difficult it is to recover the input from its representation. The benefit of compressing representations is also consistent with Occam's Razor interpretation of the ``minimum description length'' principle of \cite{mdl}.
      
        \paragraph*{Layer-wise regularization.} A DNN is composed of a number $L$ of layers that transform sequentially the input. Each one can be seen as an intermediate representation variable, noted $Y_{\la}$ for layer $\la$, that is determined by the output of the previous layer and a set of parameters $\w_{\la}$. Each layer filters out part of the information from the initial input. Thus, from the data processing inequality in \cite{element}, the following inequalities can be derived for any layer $\la$:
        \begin{equation*}
            H(Y_\la \mid C) \le H(Y_{\la-1}\mid C) \le ... \le  H(Y_{1}\mid C) \le H(X\mid C).
        \end{equation*}
        The conditional entropy of a layer impacts the conditional entropy of the subsequent layers. Consequently, similarly to the recommendation of \cite{IBdeep}, as illustrated on Fig. \ref{fig:layer-wize-regularization}, we apply a regularization on all layers, minimizing the layer-wise entropy $H(Y_\la \mid C)$, and producing a global criterion:
        \begin{equation}
            \Omega_{\mathrm{layers}} = \sum_{\la = 1}^L H(Y_\la \mid C).
        \end{equation}
     
        \paragraph*{Unit-wise regularization.} Examining one layer $\la$, its representation variable is a random vector of coordinates $Y_{\la,i}$ and of dimension $D_\la$: $Y_\la = (Y_{\la,1},..., Y_{\la, D_{\la}})$. The upper bound\footnote{This upper bound is well justified in deep learning as the neurons of a layer tend to be more and more independent of each other as we go deeper within the network.} 
        $H(Y_\la \mid C) \le \sum_{i=1}^{D_\la} H(Y_{\la,i} \mid C)$ allows us to consider the different units of a layer independently and then to define a unit-wise criterion that SHADE seeks to minimize. For each unit $i$ of every layer $\la$ we design a loss
                $\omega_{\mathrm{unit}}(Y_{\la,i} \mid C)=H(Y_{\la,i} \mid C)$ that will be part of the global regularization loss:
        \begin{align}
        \label{SHADE}
            \Omega_{\mathrm{layers}} \le \Omega_{\mathrm{units}} = \sum_{l=1}^L \sum_{i = 1}^{D_\la} \underbrace{H(Y_{\la,i} \mid C)}_{\omega_{\mathrm{unit}}(Y_{\la,i}\mid C)}.
        \end{align}
    
        In the sections that follow, we use the notation $Y$ instead of $Y_{\la,i}$ for simplicity since the coordinates are all considered independently to define $\omega_{\mathrm{unit}}(Y_{\la,i}\mid C)$.
        
    \subsection{Estimating Entropy}
        In this section, we describe how to define a loss based on the measure $H(Y\mid C)$ with $Y$ being one coordinate variable of one layer. Defining this loss is not obvious as the gradient of $H(Y\mid C)$ with respect to the layer's parameters may be computationally intractable. $Y$ has an unknown distribution and without modeling it properly it is not possible to compute $H(Y\mid C)$ precisely for the following reasons.
         
        Since $H(Y\mid C) = \sum_{c \in \C}p(c)H(Y\mid c)$ most of the estimators require to compute $|\C|$ different entropies $H(Y\mid c)$. This means that, given a batch, the number of samples used to estimate one of these entropies is divided by $|\C|$ on average which becomes particularly problematic when dealing with a large number of classes such as the 1,000 classes of ImageNet. Furthermore, entropy estimators are extremely inaccurate considering the number of samples in a batch. For example, MLE estimators of entropy in \cite{entropyestimation} converge in $\mathcal{O}(\nicefrac{(\log K)^2}{K})$ for $K$ samples. Finally, most estimators such as MLE require discretizing the space in order to approximate the distribution \textit{via} a histogram. This raises issues on the bins definition considering that the variable distribution is unknown and varies during the training in addition to the fact that having a histogram for each neuron is computationally and memory consuming.
        
        To tackle these drawbacks we investigate the two following tricks: the introduction of a latent variable that enables to use more examples to estimate the entropy; and a bound on the entropy of the variable by an increasing function of its variance to avoid the issue of entropy estimation with an histogram, making the computation tractable and scalable.
    
        \paragraph*{Latent code $\bm{Z}$.}      
            Intermediate features of a DNN most likely take similar values for inputs of different classes -- this is especially true for low-level features. The semantic information provided by a single feature $Y$ therefore describes a particular pattern it detects rather than the detection of a class. Only the association of features allows determining the class. To better understand how the class information is encoded in an individual neuron variable (before ReLU), let us take a look at the behavior of the activation function used. The ReLU activation makes the neuron act as a detector, returning a signal when a certain pattern is present on the input. If the pattern is absent the signal is zero, otherwise, $Y$ quantifies the resemblance with it.
            
            We therefore propose to associate a Bernoulli variable $Z$ to each unit variable $Y$. This variable $Z$ indicates if a particular pattern is present on the input ($Z=1$ when $Y \gg 0$) or not ($Z=0$ when $Y \le 0$). It acts like a latent code in variational models \cite{aevb} or in generative models \cite{infogan}. 
            
            In other words, $Z$ is a semantically meaningful factor about the class $C$ and from which the input $X$ is generated. The feature value $Y$ is then a quantification of the possibility for this attribute to be present ($Z=1$) or not ($Z=0$) in the input. For instance, for high level features, $Z$ could represent the presence or not of a particular object, that allows to discriminate between classes (\textit{e.g.} for a wheel, presents on cars and trucks, $Y$ notifies on the resemblance with a certain pattern representing a wheel while $Z$ indicates if the wheel is present or not on the image). Note that $Z$ is not a deterministic variable of $Y$.
            
            Therefore we assume the Markov chain $C \rightarrow Z \rightarrow X \rightarrow Y$ (see definition in \cite{element}). We indeed expect $Y$ to evolve towards a sufficient statistic of $Z$ for $C$ during the training. Considering the sufficient statistic relation $I(Y, C) = I(Y, Z)$ we get the equivalent equality $H(Y\mid C)= H(Y\mid Z)$, to finally obtain:
            \begin{align*}
                \omega_{\mathrm{unit}}(Y\mid C) &= H(Y\mid C)= H(Y\mid Z) \\
                &= \sum_{z\in\{0,1\}} p(z) H(Y\mid Z=z).
            \end{align*}
            
            This modeling of $Z$ as a binomial variable (one for each unit) has the advantage of enabling good estimators of conditional entropy since we only divide the batch into two sets for the estimation ($z=0$ and $z=1$) regardless of the number of classes.

        \paragraph*{Variance bound.}
            The previous trick allows computing fewer entropy estimates to obtain the global conditional entropy, thus increasing the sample size used for each entropy estimation. Unfortunately, it does not solve the bin definition issue. To address this, we propose to use the following bound on $H(Y\mid Z)$, that does not require the definition of bins: $H(Y \mid Z) \le \frac{1}{2}\ln\big(2 \pi e \var(Y\mid Z)\big).$
            
            This bound holds for any continuous distributions $Y$ and there is equality if the distribution is Gaussian, which is a proper law to model the activations, according to \cite{selforganizingNN}. For many other distributions such as the exponential ones, the entropy is also directly equal to an increasing function of the variance. In addition, one main advantage is that variance estimators are much more robust than entropy estimators, converging in $\mathcal{O}(\nicefrac{1}{K})$ for $K$ samples instead of $\mathcal{O}(\nicefrac{\log(K)^2}{K})$.
            
            Finally, the $\ln$ function being one-to-one and increasing, we only keep the simpler term $\var(Y\mid Z)$ to design our final loss:
            \begin{equation*}
                \Omega_\mathrm{SHADE} = \sum_{\la=1}^{L}\sum_{i=1}^{D_\la}\sum_{z \in \{0,1\}} p(Z_{\la,i} = z \mid Y) \var(Y\mid Z_{\la,i}=z).
            \end{equation*}
            
            In next section, we detail the definition of the differential loss using $\var(Y\mid Z)$ as a criterion computed on a mini-batch. 
            
        \subsection{Instantiating SHADE}   
        \label{instance}
        
            \begin{algorithm}[tb]
                    \caption{Moving average updates:
                    for $z \in \{0,1 \}$, $p^z$ estimates $p(Z = z)$ and $\mu^z$ estimates $\E(Y\mid Z = z)$}
                    \label{alg:maupdate}
                    \begin{algorithmic}[1]
                    \State \textbf{Initialize:} $\mu^0 = -1$, $\mu^1 = 1$, $p^0 = p^1 = 0.5$, $\lambda=0.8$             
                    \renewcommand{\algorithmicforall}{\textbf{for each}}
                    \ForAll{mini-batch $\{y^{(k)}, k \in 1 .. K\}$}
                        \For{$z \in \{ 0,1 \}$}
                            \State $p^{z} \leftarrow  \lambda p^{z} +   (1- \lambda)\frac{1}{K}\sum_{k=1}^K p(z\mid y^{(k)})$
                            \State $\mu^{z} \leftarrow \lambda  \mu^{z} +  (1-\lambda)\frac{1}{K}\sum_{k=1}^K\displaystyle \frac{p(z\mid y^{(k)})}{p^z}y^{(k)}$
                        \EndFor
                    \EndFor
                \end{algorithmic}
            \end{algorithm}
            
            For one unit of one layer, the previous criterion writes:
            \begin{align}
                \!\!\!\var(Y \mid Z) &= \int_{\Y} p(y) \int_{\Z}p(z\mid y)\big(y-\E(Y\mid z)\big)^2 \dz \dy \\
                &\approx \frac 1 K \sum_{k=1}^K \left[\int_{\Z}\!p(z\mid y^{(k)})\big(y^{(k)}\!-\E(Y\mid z)\big)^2 \dz \right]
                \label{eq:MC}
            \end{align}
            The quantity $\var(Y \mid Z)$ can be estimated with Monte-Carlo sampling on a mini-batch of input-target pairs $\big\{(x^{(k)}, c^{(k)})\big\}_{1 \le k  \le K}$ of intermediate representations $\big\{y^{(k)}\big\}_{1 \le k \le K}$ as in Eq.~(\ref{eq:MC}).
            
            $p(Z \mid y)$ can be interpreted as the probability of presence of attribute $Z$ on the input, so it should clearly be modeled such that $p(Z = 1 \mid y)$ increases with $y$. 
            We suggest using:
            \begin{equation*}
                p(Z=1\mid y) = \sigma(y) \qquad p(Z=0 \mid y) = 1- \sigma(y) 
            \end{equation*}
            with $\sigma(y) = 1-e^{-\textrm{ReLU}(Y)}$.
            
            For the expected values $\mu^z = \E(Y\mid z)$  we use a classic moving average that is updated after each batch as described in Algorithm \ref{alg:maupdate}. Note that the expectations are not changed by the optimization since they have no influence on the entropy $H(Y\mid Z)$.
            
        
            For this proposed instantiation, our SHADE regularization penalty takes the form:
            \begin{equation*}
                \Omega_{\mathrm{SHADE}} = \sum_{\la=1}^{L}\sum_{i=1}^{D_\la}\sum_{k=1}^K \sum_{z \in \{0,1\}} p\left(Z_{\la,i} = z\,\middle|\, y_{\la,i}^{(k)}\right) 
                \left({y_{\la,i}^{(k)}} - \mu_{\la,i}^ z\right)^2.
            \end{equation*}
    
    We have presented a regularizer that is applied layer-wise and that can be integrated into the usual optimization process of a DNN. The additional computation and memory usage induced by SHADE is almost negligible (computation and storage of two moving averages per neuron). Namely, SHADE adds half as many parameters as batch normalization does. 

\section{Experiments}
\label{sec:expes}
    \subsection{Image Classification with Various Architectures on CIFAR-10}
    
        \begin{table}[h]
            \caption{Classification accuracy (\%) on CIFAR-10 test set.}
            \label{accuracies}
                \centering
                \begin{tabular}{lccccc}
                \toprule
                & MLP & AlexNet & ResNet & Inception\\
                \midrule
                No regul.        & 62.38 & 83.25 & 89.84 & 90.97 \\
                Weight decay     & 62.69 & 83.54 & 91.71 & 91.87 \\
                Dropout          & 65.37 & 85.95 & 89.94 & 91.11 \\
                \cmidrule{1-5}
                SHADE            & 66.05 & 85.45 & \textbf{92.15} & \textbf{93.28}\\
                SHADE+D& \textbf{66.12} & \textbf{86.71} &  92.03 &  92.51\\
                \bottomrule
                \end{tabular}
        \end{table}
         
        We perform image classification on the CIFAR-10 dataset, which contains 50k training images and 10k test images of 32$\times$32 RGB pixels, fairly distributed within 10 classes \cite{cifar}. Following the architectures used in \cite{rethinking}, we use a small Inception model, a three-layer MLP, and an AlexNet-like model with 3 convolutional and 2 fully connected layers. We also use a ResNet architecture from \cite{wideresnet}. Those architectures represent a large family of DNN and some have been well studied in \cite{rethinking} within the generalization scope. For training, we use randomly cropped images of size 28$\times$28 with random horizontal flips. For testing, we simply center-crop 28$\times$28 images. We use momentum SGD for optimization, same protocol as \cite{rethinking}.
           
        We compare SHADE with two regularization methods: {\em weight decay} and {\em dropout}. 
        For all architectures, the regularization parameters have been cross-validated to find the best ones for each method and the obtained accuracies on the test set are reported in Table~\ref{accuracies}.
        
        We obtain the same trends as \cite{rethinking}, which provides a small improvement of 0.31\% over weight decay on AlexNet. The improvement over weight decay is slightly more important with ResNet and Inception (0.87\% and 0.90\%) probably thanks to the use of batch normalization. In our experiments dropout improves generalization performances only for AlexNet and MLP. It is known that the use of batch normalization lowers the benefit of dropout, which is in fact not used in~\cite{resnet}.
        
        We first notice that for all kinds of architectures the use of SHADE significantly improves the generalization performance. It demonstrates the ability of SHADE to regularize the training of deep architectures.
        
        Finally, SHADE shows better performances than dropout on all architecture except AlexNet, for which they seem to be complementary, probably because of the very large number of parameters in the fully-connected layers, with best performances obtained with SHADE coupled with dropout (named SHADE+D). This association is also beneficial for MLP. On Inception and ResNet, even if dropout and SHADE independently improve generalization performances, their association is not as good as SHADE alone, probably because it enforces too much regularization.

    \subsection{Large Scale Classification on ImageNet}
    \label{imagenetExperiment}
        In order to test SHADE regularization on a very large scale dataset, we train on ImageNet \cite{ImageNet} a WELDON network from \cite{weldone} adapted from ResNet-101. This architecture changes the forward and pooling strategy by using the network in a fully-convolutional way and adding a max+min pooling, thus improving the performance of the baseline network.
        We used the \textbf{pre-trained weights of ResNet-101} (from the torchvision package of PyTorch) achieving performances on the test set of \textbf{77.56\%} for top-1 accuracy and 93.89\% for top-5 accuracy. Provided with the pre-trained weights, the \textbf{WELDON architecture} obtains \textbf{78.51\%} for top-1 accuracy and 94.65\% for top-5 accuracy. After fine tuning the network using \textbf{SHADE} for regularization we finally obtained \textbf{80.14\%} for top-1 accuracy and 95.35\% for top-5 accuracy for a concrete improvement. This demonstrates the ability to apply SHADE on very large scale image classification successfully.

    \subsection{Training with a Limited Number of Samples}
        \begin{figure}[htb]
            \centering
            (a) MNIST-M \\ \includegraphics[width=0.32\textwidth]{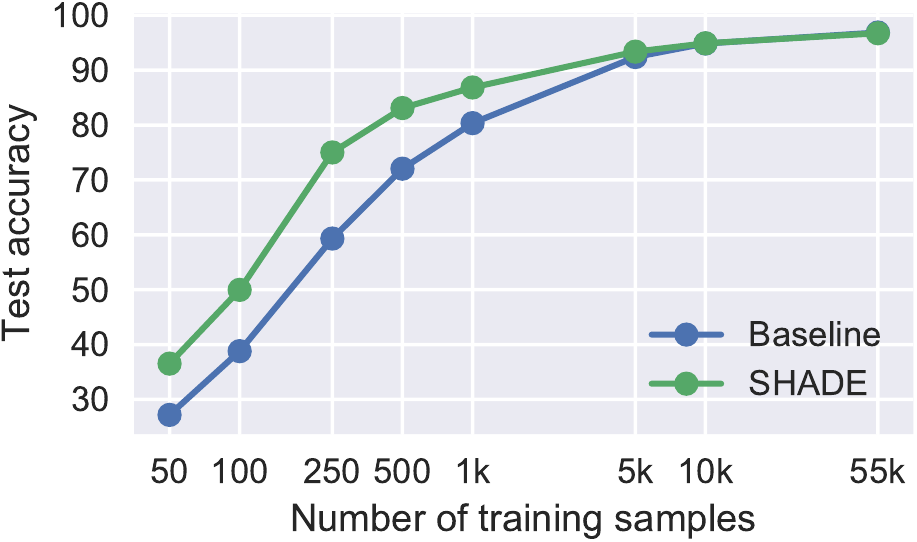} \\ 
            \vspace{1mm}
            (b) CIFAR 10 \\ \includegraphics[width=0.32\textwidth]{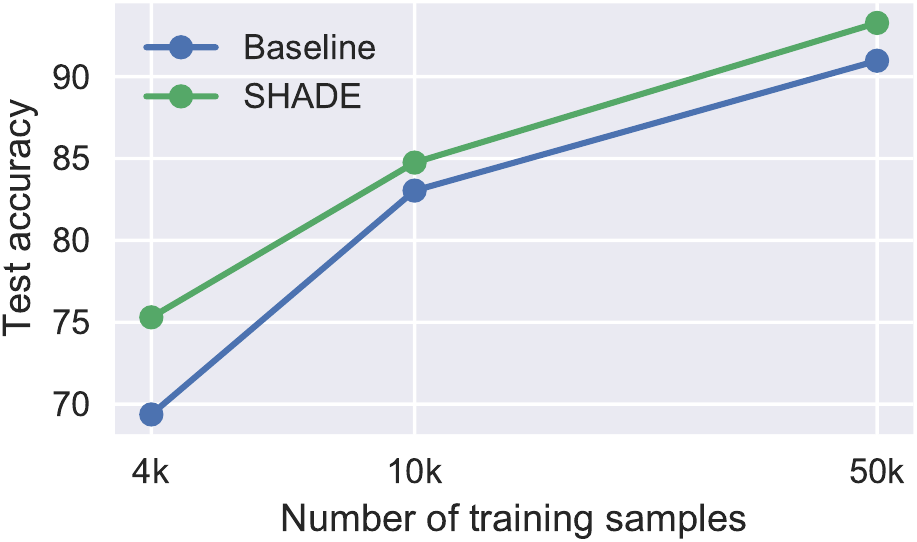}
            \caption{Results when training with a limited number of samples with and without SHADE for MNIST-M (a), and CIFAR 10 (b).}
            \label{fig:limited}
        \end{figure}
        
        \begin{figure}[htb]
            \centering
            \includegraphics[width=0.25\textwidth]{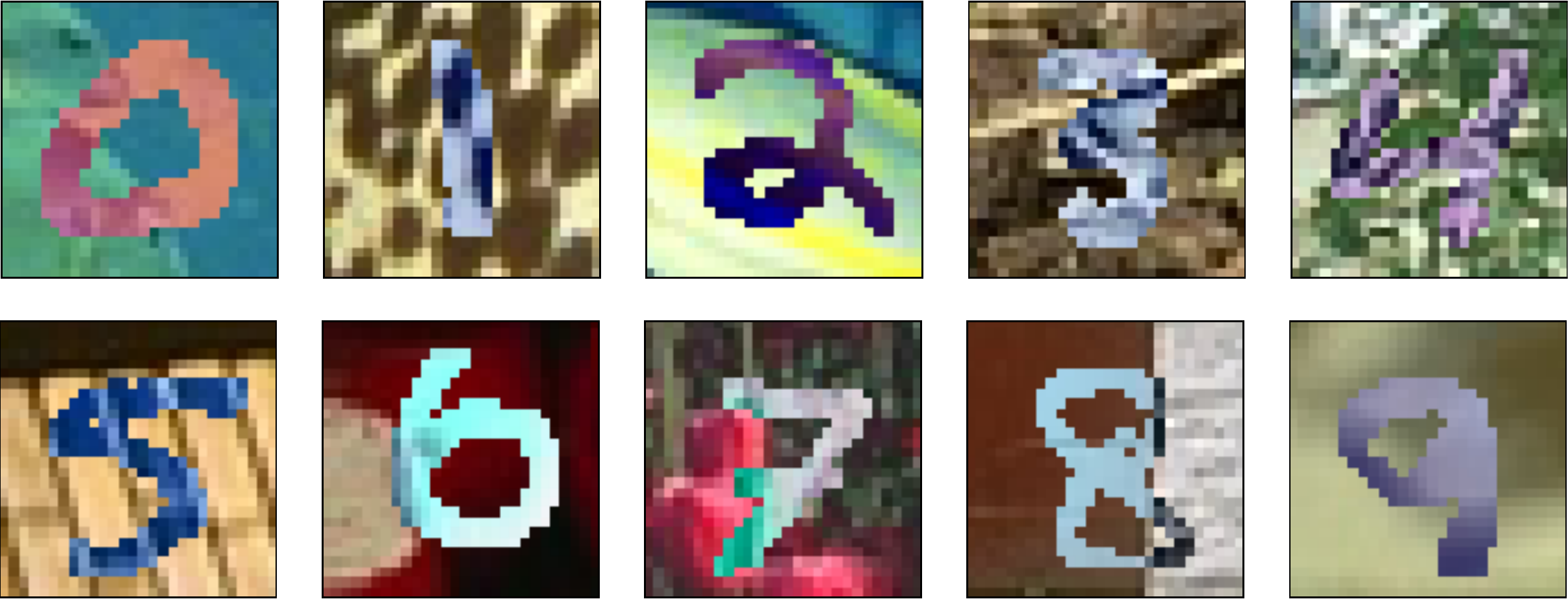}
            \caption{Examples of MNIST-M images misclassified by the baseline and correctly classified using SHADE, both trained with 250 samples.}
            \label{fig:limited_mnistm_viz}
        \end{figure}  
    
        
        When datasets are small, DNNs tend to overfit quickly and regularization becomes essential. Because it tends to filter out information and make the network more invariant, SHADE seems to be well fitted for this task. To investigate this, we propose to train DNNs with and without SHADE on 
        MNIST-M \cite{ganin2015unsupervised} with different numbers of samples in the training set.
        
        First, we tested this approach on the digits dataset MNIST-M. This dataset consists of the MNIST digits where the background and digit have been replaced by colored and textured information (see Fig. \ref{fig:limited_mnistm_viz} for examples). The interest of this dataset is that it contains lots of unnecessary information that should be filtered out, and is therefore well adapted to measure the effect of SHADE. A simple convolutional network has been trained with different numbers of samples of MNIST-M and the optimal regularization weight for SHADE has been determined on the validation set.
        The results can be seen on Figure~\ref{fig:limited}a. We can see that especially for small numbers of training samples ($<$ 1000), SHADE provides an important gain of 10 to 15\% points over the baseline. This shows that SHADE helped the model in finding invariant and discriminative patterns using less data samples.
        
        Additionally, Figure~\ref{fig:limited_mnistm_viz} shows samples that are misclassified by the baseline model but correctly classified when using SHADE. These images contain a large amount of intra-class variance (color, texture, etc.) that is not useful for the classification task, explaining why adding SHADE, that encourages the model to discard information, allows important performance gains on this dataset and especially when only few training samples are given.
        
        Finally, to confirm this behavior, we also applied the same procedure in a more conventional setting by training an Inception model on CIFAR-10. Figure~\ref{fig:limited}b shows the results in that case. We can see that once again SHADE helps the model gain in performance and that this behavior is more noticeable when the number of samples is limited, allowing a gain of 6\% when using 4000 samples.

\section{Conclusion}
    In this paper, we introduced a new regularization method for DNNs training, SHADE, which focuses on minimizing the entropy of the representation conditionally to the labels.
    This regularization aims at increasing the intra-class invariance of the model while keeping class information. 
    SHADE is tractable, adding only a small computational overhead when included into an efficient SGD training.  
    We show that our SHADE regularization method significantly outperforms standard approaches such as weight decay or dropout with various DNN architectures on CIFAR-10.
    We also validate the scalability of SHADE by applying it on ImageNet. 
    The invariance potential brought out by SHADE is further illustrated by its ability to ignore irrelevant visual information (texture, color) on MNIST-M. Finally, we also highlight the increasing benefit of our regularizer when the number of training examples becomes small.

\bibliographystyle{IEEEbib}
\bibliography{bibli}

\begin{thebibliography}{10}

\bibitem{alexnet}
Alex Krizhevsky, Ilya Sutskever, and Geoffrey~E. Hinton,
\newblock ``Imagenet classification with deep convolutional neural networks,''
\newblock in {\em NIPS}, F.~Pereira, C.~J.~C. Burges, L.~Bottou, and K.~Q.
  Weinberger, Eds., 2012.

\bibitem{resnet}
Kaiming He, Xiangyu Zhang, Shaoqing Ren, and Jian Sun,
\newblock ``Deep residual learning for image recognition,''
\newblock in {\em CVPR}, 2016.

\bibitem{weightdecay}
Anders Krogh and John~A. Hertz,
\newblock ``A simple weight decay can improve generalization,''
\newblock in {\em NIPS}, 1992.

\bibitem{dropout}
Nitish Srivastava, Geoffrey Hinton, Alex Krizhevsky, Ilya Sutskever, and Ruslan
  Salakhutdinov,
\newblock ``Dropout: A simple way to prevent neural networks from
  overfitting,''
\newblock {\em JMLR}, 2014.

\bibitem{batchnorm}
Sergey Ioffe and Christian Szegedy,
\newblock ``Batch normalization: Accelerating deep network training by reducing
  internal covariate shift,''
\newblock {\em JMLR}, 2016.

\bibitem{rethinking}
Chiyuan Zhang, Samy Bengio, Moritz Hardt, Benjamin Recht, and Oriol Vinyals,
\newblock ``Understanding deep learning requires rethinking generalization,''
\newblock {\em ICLR}, 2017.

\bibitem{scattering}
Stephane Mallat,
\newblock ``Group invariant scattering,''
\newblock in {\em Communications on Pure and Applied Mathematics}, 2012.

\bibitem{bruna}
Joan Bruna and Stephane Mallat,
\newblock ``Invariant scattering convolution networks,''
\newblock in {\em IEEE TRANSACTIONS ON PATTERN ANALYSIS AND MACHINE
  INTELLIGENCE}, 2013.

\bibitem{DBLP:journals/corr/PereyraTCKH17}
Gabriel Pereyra, George Tucker, Jan Chorowski, Lukasz Kaiser, and Geoffrey~E.
  Hinton,
\newblock ``Regularizing neural networks by penalizing confident output
  distributions,''
\newblock in {\em ICLR Workshop}, 2017.

\bibitem{IB}
N.~Tishby, F.~C. Pereira, and W.~Bialek,
\newblock ``The information bottleneck method,''
\newblock {\em Annual Allerton Conference on Communication, Control and
  Computing}, 1999.

\bibitem{element}
T.~Cover and J.~Thomas,
\newblock ``Elements of information theory,''
\newblock {\em Wiley New York}, 1991.

\bibitem{IBvariational}
A.~A. Alemi, I.~Fischer, J.~V Dillon, and K.~Murphy,
\newblock ``Deep variational information bottleneck,''
\newblock in {\em ICLR}, 2017.

\bibitem{mdl}
J.~Rissanen,
\newblock ``Modeling by shortest data description,''
\newblock {\em Automatica}, 1978.

\bibitem{IBdeep}
Tishby Naftali and Zaslavsky Noga,
\newblock ``Deep learning and the information bottleneck principle,''
\newblock in {\em Information Theory Workshop (ITW)}. IEEE, 2015.

\bibitem{entropyestimation}
Liam Paninski,
\newblock ``Estimation of entropy and mutual information,''
\newblock {\em Neural Computation}, 2003.

\bibitem{aevb}
Diederik~P. Kingma and Max Welling,
\newblock ``Auto-encoding variational bayes,''
\newblock in {\em ICLR}, 2014.

\bibitem{infogan}
Xi~Chen, Yan Duan, Rein Houthooft, John Schulman, Ilya Sutskever, and Pieter
  Abbeel,
\newblock ``Infogan: Interpretable representation learning by information
  maximizing generative adversarial nets,''
\newblock in {\em NIPS}, Jun 2016.

\bibitem{selforganizingNN}
Günter Klambauer, Thomas Unterthiner, Andreas Mayr, and Sepp Hochreiter,
\newblock ``Self-normalizing neural networks,''
\newblock in {\em Advances in Neural Information Processing Systems 30}, Guyon
  I., Luxburg~U. V., Bengio S., Wallach H., Fergus R., Vishwanathan S., and
  Garnett R., Eds., p. 972–981. Curran Associates, Inc., 2017.

\bibitem{cifar}
A.~Krizhevsky,
\newblock {\em Learning multiple layers of features from tiny images},
\newblock Ph.D. thesis, Computer Science Department University of Toronto,
  2009.

\bibitem{wideresnet}
Sergey Zagoruyko and Nikos Komodakis,
\newblock ``Wide residual networks,''
\newblock in {\em arXiv}, 2016.

\bibitem{ImageNet}
Olga Russakovsky, Jia Deng, Hao Su, Jonathan Krause, Sanjeev Satheesh, Sean Ma,
  Zhiheng Huang, Andrej Karpathy, Aditya Khosla, Michael Bernstein,
  Alexander~C. Berg, and Li~Fei-Fei,
\newblock ``Imagenet large scale visual recognition challeng,''
\newblock {\em ICJV}, 2015.

\bibitem{weldone}
Durand Thibaut, Thome Nicolas, and Cord Matthieu,
\newblock ``{WELDON: Weakly Supervised Learning of Deep Convolutional Neural
  Networks},''
\newblock in {\em CVPR}, 2016.

\bibitem{ganin2015unsupervised}
Ganin Yaroslav and Lempitsky Victor,
\newblock ``Unsupervised domain adaptation by backpropagation,''
\newblock in {\em ICML}, 2015, pp. 1180--1189.

\end{thebibliography}

\end{document}